%% file: acl2023.tex
\pdfoutput=1

\documentclass[11pt]{article}

\usepackage{ACL2023}
\usepackage{times}
\usepackage{latexsym}

\usepackage{afterpage}
\usepackage[T1]{fontenc}

\usepackage[utf8]{inputenc}

\usepackage{microtype}

\usepackage{inconsolata}

\usepackage{pifont}

\title{REAR: A Relevance-Aware Retrieval-Augmented Framework for Open-Domain Question Answering}

\author{
    \textbf{Yuhao Wang}\textsuperscript{\rm{1}}\thanks{~~Equal contributions.} \quad
    \textbf{Ruiyang Ren}\textsuperscript{\rm{1}}\footnotemark[1] \quad 
    \textbf{Junyi Li}\textsuperscript{\rm{3}} \quad 
    \textbf{Wayne Xin Zhao}\textsuperscript{\rm{1}}\thanks{~~Corresponding authors.} \\ 
    \textbf{Jing Liu}\textsuperscript{\rm{4}}\footnotemark[2] \quad 
    \textbf{Ji-Rong Wen}\textsuperscript{\rm{1,2}}
    \\
    \textsuperscript{1}Gaoling School of Artificial Intelligence, Renmin University of China\\
    \textsuperscript{2}School of Information, Renmin University of China \\
    \textsuperscript{3}Department of Computer Science, National University of Singapore \quad \textsuperscript{4}Baidu Inc.\\
    \{yh.wang500, reyon\_ren\}@outlook.com, batmanfly@gmail.com
}

\begin{document}
\maketitle
\begin{abstract}
Considering the limited internal parametric knowledge,  retrieval-augmented generation~(RAG) has been widely used to extend the knowledge scope of large language models~(LLMs).  
Despite the extensive efforts on RAG research, in existing methods, LLMs cannot precisely assess the relevance of retrieved documents, thus likely leading to misleading or even incorrect utilization of external knowledge (\ie retrieved documents). To address this issue, in this paper, we propose  \textbf{REAR}, a \textbf{RE}levance-\textbf{A}ware \textbf{R}etrieval-augmented approach for open-domain question answering~(QA). As the key motivation, we aim to enhance the self-awareness regarding the reliability of external knowledge for LLMs, so as to adaptively utilize external knowledge in RAG systems. 
Specially, we develop a novel architecture for LLM-based RAG systems, by incorporating a specially designed assessment module that precisely assesses the relevance of retrieved documents. Furthermore, we propose an improved training method based on bi-granularity relevance fusion and noise-resistant training. By combining the improvements in both architecture and training, our proposed REAR can better utilize external knowledge by effectively perceiving the relevance of retrieved documents. Experiments on four open-domain QA tasks show that REAR significantly
outperforms previous a number of competitive RAG approaches. 
Our codes can be accessed at \url{https://github.com/RUCAIBox/REAR}.
\end{abstract}

\input{sec/intro}
\input{sec/related_work}
\input{sec/preliminaries}
\input{sec/methodology}
\input{sec/experiments}
\input{sec/results}

\section{Conclusion}
In this paper, we aimed to enhance the self-awareness of source relevance in RAG systems, and proposed \textbf{REAR}, a \textbf{RE}levance-\textbf{A}ware \textbf{R}etrieval-augmented approach for open-domain question answering~(QA). 
For model architecture, we explicitly integrate an assessment module to precisely capture the relevance signals, and employ it to guide the utilization of external knowledge. For model training, we designed an improved training method with bi-granularity relevance fusion and noise-resistant training, which enhance the capacities of fine-grained relevance assessment and adaptive use of retrieved documents. Our data construction strategy collects high-quality data without access to GPT APIs. Extensive experiments on four datasets demonstrate the effectiveness and generalization of REAR's knowledge utilization.

As future work, we will extend the proposed approach REAR to deal with more fine-grained source utilization (\eg passage or sentence level augmentation), and also consider applying REAR to other knowledge-intensive tasks.  

\section*{Limitations}
For LLMs, the challenge of being misled by irrelevant retrieved documents is a significant obstacle, underscoring the crucial need for enhancing LLMs' ability to adaptively utilize retrieved documents. 
In response to this issue, our work has concentrated on refining the architecture and training methods to bolster the effective use of retrieved documents by LLMs. We have implemented document-level relevance assessment and dynamic utilization strategies, significantly boosting the factual accuracy of generated content by LLMs. However, our current approach has not delved into guiding LLMs to focus more granularly on key sentences or tokens within the retrieved documents.

Moreover, the applicability of our methods across a broader spectrum of RAG tasks, such as those encompassed by the KILT benchmark, remains to be thoroughly evaluated. This gap presents a pivotal area for our future investigations.

\bibliography{anthology,custom}
\bibliographystyle{acl_natbib}

\input{sec/appendix}

\end{document}

%% file: sec/intro.tex
\section{Introduction}

Despite the progressive capacities,  
large language models (LLMs)~\cite{brown2020language, zhao2023survey} still struggle with knowledge-intensive tasks like open-domain question answering~(QA), lacking in  real-time and domain  knowledge~\cite{li2023halueval, cheng2024small}.
To mitigate this issue, retrieval-augmented generation~(RAG) provides  LLMs with potentially relevant documents through a retrieval module~\cite{gao2023retrieval}, aiding in {generating more precise content}.

\begin{figure}
\centering
\includegraphics[width=0.44\textwidth]{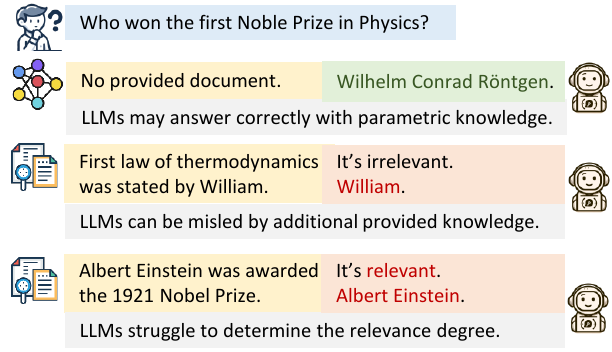}
    \caption{LLMs may be misled by irrelevant documents, and struggle to determine the relevance of a document~\cite{ren2023investigating, zhang2024large}.}
    \label{fig:documents}
\end{figure}

While RAG offers clear benefits, it also introduces several technical challenges for effectively improving LLMs. 
Firstly, the retrieved results likely contain irrelevant content or documents, which may mislead LLMs and even cause them to respond incorrectly~\cite{ren2023investigating, mallen2023not}. 
Moreover, it has become common to incorporate multiple reference documents to boost the overall reliability of retrieved documents. However, this approach potentially amplifies the impact of the noise present in the retrieved documents~\cite{liu2023lost, shi2023large}. 
Thus, LLMs face difficulties in filtering irrelevant documents and integrating their internal knowledge~\cite{dong2023bamboo}, which needs to avoid potential interference with noisy content.

Recently, several studies~\cite{luo2023sail, asai2023self, yoran2023making} have attempted to enhance the robustness of RAG systems. For instance, 
Self-RAG~\cite{asai2023self} allows the model to introspect its outputs by generating special tokens to discriminate if the documents are relevant, and RobustLM~\cite{yoran2023making}
prompts LLMs to first discriminate if the documents are relevant and then generate answers.
However, these approaches perform the assessment of document relevance solely based on binary labels, which are highly sparse and not precise to capture the fine-grained relevance. In addition, they seldom consider the varied relevance degree of reference documents, making the utilization of external knowledge somehow blind.

To this end, in this paper, we propose  \textbf{REAR}, a \textbf{RE}levance-\textbf{A}ware \textbf{R}etrieval-augmented generation approach for open-domain question answering~(QA). 
Our key idea is to develop robust self-awareness regarding the reliability of external knowledge (\ie retrieved documents) within RAG systems, so that the LLM can learn to adaptively utilize the internal and external knowledge for solving complex QA tasks. To achieve this goal, we make two major contributions in both model architecture and training. 
{First, we propose relevance-aware RAG architecture by incorporating explicit assessment modules in LLMs' generation architecture to perform an additional relevance assessment task. In our architecture, the assessment module effectively captures relevance signals, and feeds them back to avoid distractions from irrelevant external knowledge during generation. 
Secondly, to support the relevance-aware RAG architecture, we further propose two training strategies. Bi-granularity relevance fusion strategy further integrates both coarse and fine-grained relevance supervision to overcome the limitations of binary discriminative methods, while noise-resistant training strategy enhances the discrimination ability of the LLM by incorporating negatives in the training procedure. 
}
 
To the best of our knowledge, we are the first to introduce the idea of incorporating explicit assessment modules in the generation architecture of LLMs to aid in irrelevance-resistant generation. 
Extensive experiments on public open-domain QA benchmarks attest to the effectiveness of our REAR framework.
Notably, we also demonstrate the strong generalization capability of REAR by conducting out-of-domain evaluation on multiple open-domain QA benchmarks.

%% file: sec/related_work.tex
\begin{figure*}[ht!]
  \centering
  \includegraphics[width=\textwidth]{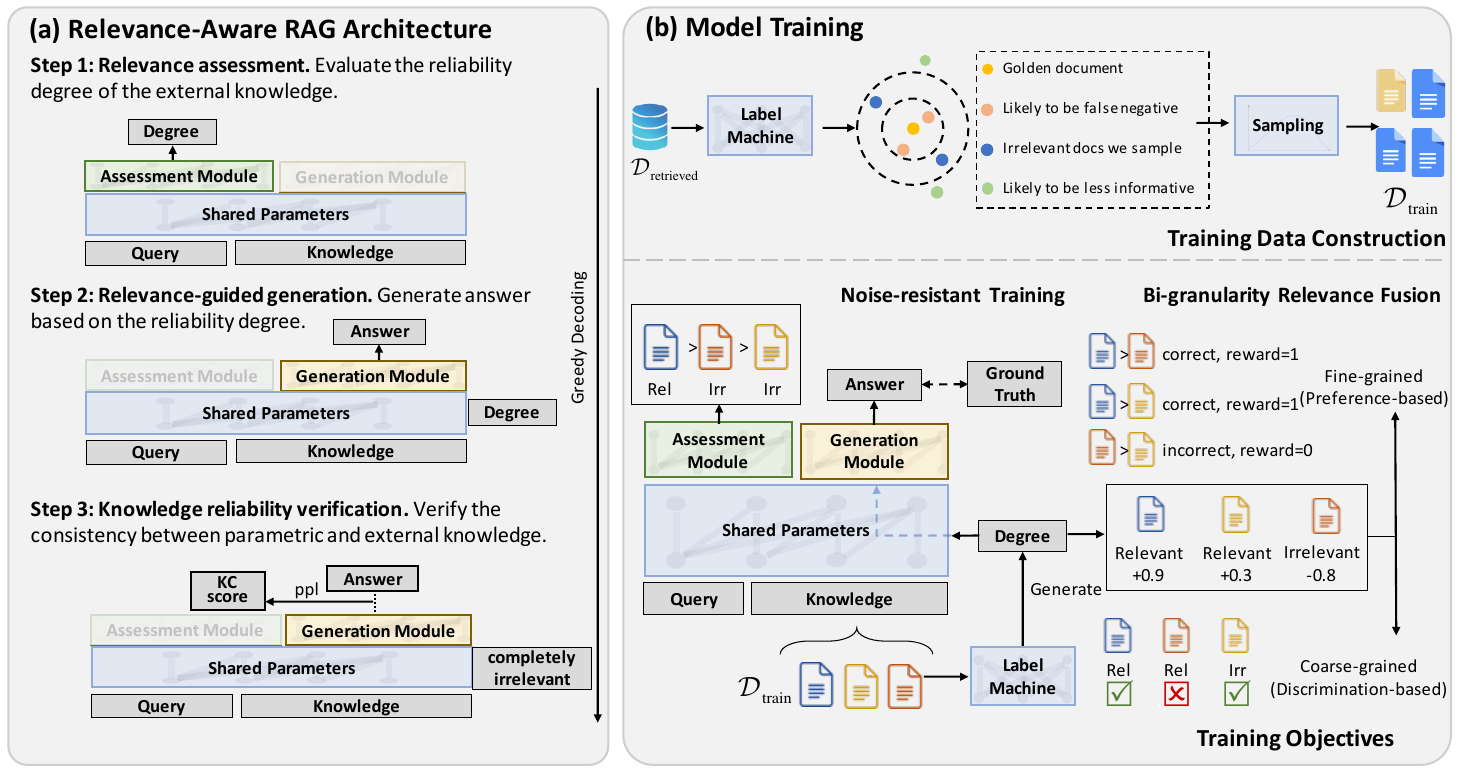}
  \caption{The overview of the proposed REAR framework.}
  \label{fig:framework}
\end{figure*}
\section{Related Work}

\paratitle{Open-domain Question Answering.}
Modern open-domain QA systems combine traditional IR techniques with neural reading comprehension models ~\cite{chen2017reading}.
After retrieving documents~\cite{ren2021pair, zhang2021adversarial}, an extractive or generative reader is typically used for answer generation~\cite{zhu2021retrieving}.
Models like 
REALM~\cite{guu2020retrieval}, 
RAG ~\cite{lewis2020retrieval}, RETRO ~\cite{borgeaud2022improving} and In-context RALM~\cite{ram2023context} have demonstrated improved factual generation capabilities. However, these readers make generation quality more prone to noise impact, for lacking explicit relevance discernment.
We propose an architecture that explicitly generates relevance scores to assist in subsequent generation tasks.

\paratitle{Retrieval-augmented LLMs.}
Several research aims at aligning the retriever outputs with the preferences of the LLMs~\cite{izacard2020distilling, sachan2021end}. 
And works like Atlas~\cite{izacard2022few}, RA-DIT~\cite{lin2023ra} jointly train the language model and the retriever for advanced performance on RAG.
Some other work improves the quality of retrieved documents by expanding the knowledge sources~\cite{li2023web} or query rewriting~\cite{zheng2023take}.
However, we focus on a scenario where the irrelevant documents from retrieval could mislead LLMs.
Several recent studies~\cite{luo2023sail, asai2023self, yoran2023making} attempt to adopt a paradigm in which an initial judgment on relevance is made by generating a statement or special token before proceeding to content generation.
However, these methods still lack accuracy in relevance discrimination and LLMs are still vulnerable to irrelevant document interference. 
Therefore, we propose a framework that can accurately assess the relevance degree, and is more robust to irrelevant content.

%% file: sec/preliminaries.tex
\section{Task Formulation}

In this work, we focus on the task of open-domain question answering~(QA)~\cite{chen2017reading, zhao2023dense}, aiming at answering questions using a large collection of documents.
Typically, open-domain QA tasks are often tackled with a  \emph{retriever-reader} approach~\cite{chen2020open}, where the retriever finds relevant evidence and the reader generates the answer based on the retrieved evidence. 

Formally, given a query $q$, the retriever outputs top-$k$ documents $\mathcal{D}=\{ d_i \}_{i=1}^k$ from a document collection (can be refined by an optional \emph{reranker}) at the first stage. 
Different from prior studies that combine the entire set of retrieved documents as a unified reference for answer generation~\cite{luo2023sail, xu2023recomp, hofstatter2023fid}, our approach emphasizes individual document utilization, which can be also extended to a multi-document setting.
Given the input query $q$ and reference documents $\mathcal{D}=\{ d_i \}_{i=1}^k$, the reader (\ie the LLM) generates an answer $a_i$ based on each reference document 
$d_i$, forming an answer set $\mathcal{A}$:
\begin{equation}
    \mathcal{A} = \{a_i\}_{i=1}^k = \{\text{LLM}(q, d_i) \mid d_i \in \mathcal{D}\}.
\end{equation}
Subsequently, we can choose the final answer from $\mathcal{A}$ based on some specific ways, ensuring it aligns best with the query $q$.

Based on this task formulation, we consider enhancing two key aspects: precise evaluation of relevance between queries and documents (\emph{identifying relevant references}), and leveraging relevance signal for noise-resistant generation (\emph{reducing the influence of irrelevant content}). Therefore, we introduce a relevance-aware approach designed specifically for these challenges.

%% file: sec/methodology.tex
\section{Methodology}

In this section, we present the proposed \textbf{RE}levance-\textbf{A}ware \textbf{R}etrieval-augmented generation framework (\textbf{REAR}), 
which is capable of precisely assessing the relevance degree during the generation process by incorporating explicit assessment modules within the LLM.
Furthermore, we propose optimized training methods that are compatible with the REAR framework to support efficient operation, including bi-granularity relevance fusion and noise-resistant training.

\subsection{Relevance-Aware RAG Architecture}
\label{subsec:architecture}
In this part, we propose a novel architecture
that augments the LLM with a relevance-assessing module for enhancing the awareness of irrelevant interference.
As shown in Fig.~\ref{fig:framework}~(a), the inference of the REAR architecture encompasses three steps, {including relevance assessment, relevance-guided generation, and knowledge reliability verification.}

\subsubsection{Relevance Assessment}
\label{subsec:relass}
Instead of treating all the retrieved documents equally, we first aim to assess the relevance degrees of the documents.
Drawing from the success of LLM-based decoder in achieving precise relevance assessment~\cite{Sun2023IsCG, ma2023fine}, we first map the input query-document pair into the relevance embedding $\bm{v}_\text{rel}$ by the LLM:
\begin{equation}
    \label{eq:dense_rep}
    \bm{v}_\text{rel} = \text{LLM}(q, d)[-1].
\end{equation}
Subsequently, $\bm{v}_\text{rel}$ is quantified into a score $s_\text{rel}$ by the assessment module:
\begin{equation}
    \label{eq:score_gen}
    s_{\text{rel}} = \text{Assess}(\bm{v}_\text{rel}),
\end{equation}
where $\text{Assess}(\cdot)$ is the assessment module implemented as a linear projection layer. 

\subsubsection{Relevance-guided Generation}

Different from previous works that ignore the relevance of document~\cite{cuconasu2024power}, we aim to integrate the relevance score of each document into LLMs to assess document reliabilities and subsequently guide the generation process. 
Since the relevance score $s_\text{rel}$ (in Eq.~\ref{eq:score_gen}) is a scalar, which may not be fully utilized by LLMs, we further incorporate  an embedding layer to map it into a dense vector $\bm{v}_\text{guide}$ as: 
\begin{equation}
    \bm{v}_\text{guide} = \text{Embedding}(s_\text{rel}). 
\label{eq:gen_emb}
\end{equation}
This embedding vector serves as a cue for the LLM to generate an answer $a$ based on either the internal knowledge of LLM (the relevance score is low) or external evidence (the relevance score is high) as:  
\begin{equation}
    a = \text{LLM}(q, d, \bm{v}_\text{guide}). 
\label{eq:ans_gen}
\end{equation}

\subsubsection{Knowledge Reliability Verification}
Based on the generated answers,
{we finally verify the correctness of the answers by considering two factors: (a) Is the provided document reliable enough to trust the corresponding answer? (b) Without referring to the documents, to what degree will the LLM adhere to its original response?} 
Specially, we propose two strategies, namely source-reliability and knowledge-consistency.

\textbullet~\emph{Source-reliability}:
This strategy primarily emphasizes the quality of external knowledge. If an LLM assigns a high relevance score to a document, then the answer derived from it is considered more reliable.

\textbullet~\emph{Knowledge-consistency }: 
{
This approach further verifies if the provided knowledge conflicts with the parametric knowledge.}
Specifically, inspired by the success of self-consistency in Chain-of-Thought reasoning~\cite{wei2022chain,wang2022self},
we inform the LLM that the document is irrelevant by setting the relevance score to zero (denoted by \( \hat{s}_{\text{rel}} \)) and calculate the inverse of perplexity $c$~\cite{meister2021language} of generating the answer $a$:
\begin{equation}
c = \frac{1}{\text{PPL}(a \mid q, d, \hat{s}_{\text{rel}}=0)},
\label{eq:consistency}
\end{equation}
{
which evaluates the extent of LLM to stand by its original answer based on the parametric knowledge.}
Then, we linearly combine the knowledge-consistency score \(c_i\) with the relevance score \(s_{\text{rel}}(q, d_i)\) to select the final answer.

\subsection{Model Training}
\label{subsec:training}

In this part, we will introduce the training pipeline for optimizing our approach, As shown in Fig.~\ref{fig:framework}~(b).

\subsubsection{Bi-granularity
Relevance Fusion} 
\label{subsec:fcfo}
Precise relevance assessment is crucial for the reliable utilization of retrieved documents.  
Previous work often adopts the coarse-grained binary discrimination task~\cite{yoran2023making}
, which cannot provide sufficient evidence for solving complex QA tasks. 
Therefore, we consider further incorporating a preference-based fine-grained task. 
Specifically, for the fine-grained supervision, we utilize the estimated relevance scores (See Section~\ref{subsec:data-inf}) for deriving relevance
preference constraints:
\begin{equation}
\mathcal{L_\text{fine}} = -\sum_{i} \sum_{j} (s_i > s_j)\log\left(\sigma_i-\sigma_j\right),
\label{eq:fine}
\end{equation}
where $\sigma_i$ denotes the normalized probability of assessing $(q, d)$ as relevant by the LLM.
Furthermore, we combine it with the coarse-grained binary loss $\mathcal{L_\text{coarse}}$, as the objective of the bi-granularity relevance fusion: 
\begin{equation}
    \mathcal{L_\text{bi-granularity}} = \mathcal{L_\text{coarse}} + \mathcal{L_\text{fine}}.
\label{eq:fcfo}
\end{equation}

\subsubsection{Noise-resistant Training}

In addition to improving the capability of identifying relevant documents, we further consider enhancing the discrimination ability when reference documents contain irrelevant content or even noise, such that the LLM can adaptively use external evidence for task solving. 
Specially, we further incorporate negative example documents $\mathcal{D}^-$ into the original corpus $\mathcal{D}$ for optimizing LLMs:

\begin{equation}
\mathcal{L}_\text{noise-resistant} = \sum_{d \in \mathcal{D} \cup \mathcal{D}^-} \log P(a \mid q, d, s_{\text{rel}}).
\label{eq:sft}
\end{equation}

Through noise-resistant training, the LLM can learn to discern the incorporation of irrelevant documents, without being encumbered by extraneous information. 

\subsubsection{Training Data Construction}
\label{subsec:data-inf}
To optimize our model, we need high-quality training samples (both positive and negative samples) and labels. 

\paratitle{Relevance Labels Acquisition.} 
To obtain fine-grained relevance labels used in Section~\ref{subsec:fcfo}, we
employ a small-scale reranker to acquire the continuous relevance score $s_\text{ce}$.
We adopt rerankers with the cross-encoder architecture, since they are regarded as effective for assessing relevance degree~\cite{zhao2023dense, khattab2020colbert}.
In combination with the traditional
method of binary annotating label $y$, the estimated score is given as: 
\begin{equation}
    s_\text{rel} = \frac{1}{2}\left(s_{\text{ce}} + y\right).
    \label{eq:rel_score_gen}
\end{equation}
{This labeling approach combines lexical and semantic similarity, allowing for the acquisition of high-quality labels without accessing GPT APIs.}

\paratitle{Irrelevant Documents Sampling.}
The training method necessitates the use of irrelevant (negative) documents.
It has been shown that negative sampling has a large impact on relevance assessment~\cite{xiong2020approximate}. 
Specially, as shown in Fig.~\ref{fig:framework}~(b), we refine  SimANS~\cite{zhou2022simans} that ensures negatives are neither too difficult (false negatives) nor too trivial (uninformative):  
\begin{equation}
p_i \propto 
\begin{cases} 
\exp(-a(s_i - \hat{s}^+ - b)^2),
& s_i < \hat{s}^+ - b, \\
\exp(-a k (s_i - \hat{s}^+ - b)^2),& s_i \geq \hat{s}^+ - b,
\end{cases}
\label{eq:hard_neg}
\end{equation}
where the sampling probability for the hard negative document is $p_i$, $s_i$ and $\hat{s}^+$ respectively denote the relevance scores of document $d_i$ and the positive document, and $a$, $b$, and $k$ are hyperparameters. 
By incorporating a decay scaler $k$ into the sampling probability when relevance scores are high, we reduce the chance of sampling false negatives.

Finally, we define the overall loss function for our REAR framework by combining the bi-granularity loss by Eq.~\ref{eq:fcfo} and noise-resistant loss by Eq.~\ref{eq:sft}:
\begin{equation}
\mathcal{L}_\text{REAR} = \mathcal{L}_\text{bi-granularity} + \mathcal{L}_\text{noise-resistant}.
\label{eq:loss}
\end{equation}

\subsection{Discussion}
\label{subsec:discuss}

\begin{table}[t!]
\centering
\small
\scalebox{0.95}{\begin{tabular}{@{}lcccc@{}}
\toprule
\textbf{Aspect} & \textbf{Self-RAG} & \textbf{CoN} & \textbf{SAIL} & \textbf{REAR (ours)}\\
\midrule
\textbf{Assess} & Gen & Gen & Gen & Explicit Module\\
\textbf{Train} & SFT & SFT & SFT & SFT+~Contrastive Loss\\
\textbf{Data} & GPT & GPT & GPT & Free Model (110M)\\
\bottomrule
\end{tabular}}
\caption{The difference between REAR and previous work. Assess, Train and Data are short for relevance assessment method, training loss, and data construction methods respectively. REAR utilizes an explicit module for relevance assessment, and adopts bi-granularity~(involving contrastive loss) for training. Furthermore, we label the data without access to GPT APIs.
}
\label{tab:methods}
\end{table}

\paratitle{Distinctions from Existing Methods.}
{
As shown in Table~\ref{tab:methods}, our primary contribution lies in the architecture design, which differs significantly from existing studies.
Under our optimized architecture, LLMs can generate more fine-grained relevance signals to aid in the following generation process. Besides, LLMs can further calculate the consistency between parametric and external knowledge to evaluate the reliability of answers.
Moreover, this architecture makes it easy to adopt the proposed preference-based and noise-resistant loss functions. Furthermore, our label machine makes good use of smaller models and traditional labels, and our sampling strategy improves training data quality, eliminating the need for GPT APIs. As a result, REAR achieves more precise relevance evaluation and better generation performance (Table~\ref{tab:overall}). 
}

\begin{table}[t!]
\centering
\renewcommand\tabcolsep{4.4pt}
\small
\begin{tabular}{@{}lccc@{}}
\toprule
\textbf{Methods} & \textbf{T.C.} & \textbf{Training} & \textbf{Inference}\\
\midrule
CoN & $\mathcal{O}((p+nd)^2)$ & 10.34s/step & 0.82s/sample\\
Self-RAG & $\mathcal{O}(n(p+d)^2)$ & 6.52s/step & 1.41s/sample\\
\midrule
REAR (ours) & $\mathcal{O}(n(p+d)^2)$ & 6.33s/step & 0.45s/sample\\
\bottomrule
\end{tabular}
\caption{The efficiency analysis of REAR and previous work. T.C. is short for time complexity. $d$, $p$ and $n$ denote the length of the document, the length of the prompt, and the number of documents respectively.
}
\label{tab:effciency}
\end{table}

\paratitle{Efficiency.}
{
We further discuss the efficiency of our REAR, as shown in Table~\ref{tab:effciency}. First, we compare REAR with other RAG frameworks that employ different task formulations, such as Chain-of-Note (CoN)~\cite{yu2023chain}. CoN processes extensive paragraphs and generates in-depth analyses to identify usable parts of document collections. This methodology leads to increased training and inference times due to the quadratic time complexity associated with transformers~\cite{, dong2024exploring}, where time is proportional to the square of the input sequence length.
Besides, compared to Self-RAG, which follows a similar approach, REAR achieves a reduction in inference time. This improvement is primarily due to our integration of PagedAttention~\cite{kwon2023efficient}. By using PagedAttention, we ensure that calculations performed during the relevance assessment phase are preserved, thereby eliminating the need for redundant recalculations. The comparisons of actual training and inference times in Table~\ref{tab:effciency} further illustrate the computational efficiency of our method.}

%% file: sec/experiments.tex
\section{Experiments}
\label{sec:exp}
In this section, we detail the experimental setup and then report the main findings of our results.
\subsection{Experimental Setup}

\paratitle{Datasets.} 
We collect the training data from the  Natural Questions (NQ)~\cite{kwiatkowski2019natural} training set.
To ensure the model’s adaptability, we also test its performance on three additional open-domain datasets, including TriviaQA~\cite{joshi2017triviaqa}, WebQuestions (WebQ)~\cite{berant2013semantic}, and SQuAD~\cite{rajpurkar2016squad}, showing its generalization capabilities to out-of-domain data. We follow the test split in prior work~\cite{karpukhin2020dense}. The details are in Appendix~\ref{app:data}.

\begin{table*}[ht!]
\centering
\small
\begin{tabular}{lcccccccccc}
    \toprule
     \multirow{3}{*}{\textbf{LLMs}} & \multicolumn{2}{c}{\textbf{NQ}} & \multicolumn{2}{c}{\textbf{TriviaQA}} & \multicolumn{2}{c}{\textbf{WebQ}} & \multicolumn{2}{c}{\textbf{SQuAD}} & \multicolumn{2}{c}{\textbf{Average}}\\
     \cmidrule(lr){2-3} \cmidrule(lr){4-5} \cmidrule(lr){6-7} \cmidrule(lr){8-9} \cmidrule(lr){10-11} 
      & \textbf{EM} & \textbf{F1} & \textbf{EM} & \textbf{F1} & \textbf{EM} & \textbf{F1} & \textbf{EM} & \textbf{F1} & \textbf{EM} & \textbf{F1}\\
    \midrule
    \multicolumn{11}{c}{\emph{Direct Retrieval-Augmented QA}} \\
      LLaMA2-Chat$_{~\text{7B}}$ & 30.47 & 41.39 & 53.92 & 62.70 & 22.79 & 38.29 & 21.09 & 31.67 & 32.07 & 43.51
\\
      Mistral-It$_{~\text{7B}}$ & 10.83 & 31.77 & 44.59 & 62.55 & 8.71 & 30.79 & 13.78 & 34.25 & 19.48 & 39.84
\\
      Baichuan2-Chat$_{~\text{7B}}$ & 33.49 & 45.61 & 61.17 & 69.98 & 23.87 & 40.78 & 26.55 & 38.97 & 36.27 & 48.84
\\
      ChatGLM3$_{~\text{6B}}$ & 13.27 & 20.48 & 24.57 & 33.76 & 5.61 & 18.38 & 8.31 & 15.98 & 12.94 & 22.15
\\
    \midrule
    \multicolumn{11}{c}{\emph{RobustLM prompting (4-shot)}} \\
      LLaMA2-Chat$_{~\text{7B}}$ & 30.53 & 42.57 & 53.27 & 63.52 & 21.01 & 38.29 & 21.83 & 33.45 & 31.66 & 44.46
\\
      Mistral-It$_{~\text{7B}}$ & 19.11 & 32.80 & 48.31 & 59.87 & 13.63 & 30.76 & 15.98 & 28.28 & 24.26 & 37.93
\\
      Baichuan2-Chat$_{~\text{7B}}$ & 27.42 & 39.72 & 52.07 & 62.27 & 18.90 & 36.13 & 19.24 & 30.92 & 29.41 & 42.26
\\
      ChatGLM3$_{~\text{6B}}$ & 24.65 & 32.67 & 46.57 & 54.23 & 20.37 & 34.60 & 18.71 & 25.90 & 27.58 & 36.85
\\
     \midrule
    \multicolumn{11}{c}{\emph{Fine-tuned RALMs}} \\
    Self-RAG$_{~\text{7B}}$$^\dagger$ & 41.02 & 46.78 & 52.38 & 39.15 & 31.40 & 26.41 & 35.28 & 19.33 & 40.02 & 32.92
\\
    RobustLM$_{~\text{7B}}$ & 44.40 & 53.08 & 62.86 & 70.88 & 32.48 & 46.89 & 27.52 & 36.75 & 41.82 & 51.90
\\
    \textbf{REAR}$_{~\text{7B}}$ w/ Source Rel. & \uline{51.33} & \textbf{60.53} & \uline{65.36} & \uline{74.14} & \uline{33.02} & \uline{47.67}& 
\uline{36.78} & \uline{46.64} & \uline{46.62} & \uline{57.25}\\
    \textbf{REAR}$_{~\text{7B}}$ w/ Knowledge Con. & \textbf{51.41} & \uline{60.50} & \textbf{66.26} & \textbf{74.87} & \textbf{33.51} & \textbf{48.14} & \textbf{37.21} & \textbf{47.19} & \textbf{47.10} & \textbf{57.68}\\
    \bottomrule
\end{tabular}
\caption{A comparison between REAR and baselines on NQ, TriviaQA, WebQ and SQuAD datasets and the averaged performance. our REAR approach surpasses all the other baselines in QA performance.
The best and second-best results are in \textbf{bold} and \uline{underlined} fonts respectively.
Self-RAG$^\dagger$ is evaluated using accuracy (Acc) instead of EM, which is a less strict metric that measures whether the responses contain the answers.
The last two lines are our REAR with different verification strategies: ``w/~Source Rel.'' means the source-reliability strategy, and ``w/~Knowledge Con.'' means the knowledge-consistency strategy.
} 
\label{tab:overall}
\end{table*}

\paratitle{Baselines.} We consider the following two lines of baselines for comparison. 

\textbf{(1) Retrieval augmentation based prompt methods}: we design different prompting strategies based on open-source LLMs (without tuning tailored to RAG tasks) to support RAG, including 

\textbullet~\emph{Direct Retrieval-Augmented QA}: We concatenate the top 10 retrieved documents as a single reference document for RAG. To enhance EM metric accuracy, we further incorporate several answer examples within the prompts, as illustrated in Fig.~\ref{fig:instruction-format}.

\textbullet~\emph{RobustLM prompting}: We following the approach of the prompting strategy~\cite{yoran2023making}. The LLMs are required to determine document relevance before generating responses. It employs 4-shot demonstrations  (Fig.~\ref{fig:instruction-format-judgment}), and provides the top 1 retrieved document.


For open-source LLMs, we consider 
LLaMA2-Chat~\cite{touvron2023llama}, Mistral-It~\cite{jiang2023mistral}, Baichuan2-Chat~\cite{yang2023baichuan}, and ChatGLM3~\cite{du2022glm}. 

\textbf{(2) Specially designed RAG methods}: we also consider fine-tuned RobustLM~\cite{yoran2023making} and Self-RAG~\cite{asai2023self} as baselines, which have been specially optimized for the RAG tasks.  
To ensure a fair comparison, the two frameworks above are evaluated with the same set of retrieved documents as used for REAR. 

\paratitle{Metrics.}
We employ three metrics to evaluate the model's capability of QA accuracy. Exact match (\textbf{EM})~\cite{lee2019latent} and \textbf{F1} are widely adopted for open-domain QA evaluation. EM calculates whether responses exactly match the gold truth answers and calculates the precision-recall overlap of predicted and true answers.
We further evaluate the accuracy in determining the relevance of the given document for LLMs with another two metrics. 
\textbf{Hit@1} evaluates if the document referenced for the model's final answer generation is relevant. 
\textbf{JAcc}, short for judgmental accuracy, measures the proportion of documents correctly evaluated by LLMs as relevant or not.

\paratitle{Implementation Details.}
To implement our  REAR approach, we fine-tune LLaMA2-Base$_\text{7B}$~\cite{touvron2023llama} on the NQ training set for 1 epoch.
We set the learning rate to 1e-6.
For evaluation, all retrieval documents are sourced from the top 10 documents retrieved by dense retrievers (detailed in Appendix ~\ref{app:rere}).

\begin{table*}[t!]
\centering
\small
\begin{tabular}{lcccccccc}
    \toprule
     \multirow{3}{*}{\textbf{LLaMA2$_\text{7B}$}} & \multicolumn{2}{c}{\textbf{NQ}} & \multicolumn{2}{c}{\textbf{TriviaQA}} & \multicolumn{2}{c}{\textbf{WebQ}} & \multicolumn{2}{c}{\textbf{SQuAD}}\\
     \cmidrule(lr){2-3} \cmidrule(lr){4-5} \cmidrule(lr){6-7} \cmidrule(lr){8-9} 
      & \textbf{JAcc} & \textbf{Hit@1} & \textbf{JAcc} & \textbf{Hit@1}& \textbf{JAcc} & \textbf{Hit@1}& \textbf{JAcc} & \textbf{Hit@1}\\
    \midrule
       + RobustLM-prompting~\cite{yoran2023making} & 25.04 & - & 43.36 & - & 28.84 & - & 16.84 & -\\
       + RobustLM-training~\cite{yoran2023making}& 56.59 & - & 56.09 & - & 49.61 & - & 56.99 & -\\
       + Self-RAG~\cite{asai2023self} & 19.81 & 51.11 & 35.69 & 64.47 & 25.69 & 47.98 & 10.73 & 38.73\\
       + REAR~(ours) & \textbf{74.04} & \textbf{66.79} & \textbf{80.79} & \textbf{74.98} & \textbf{65.99} & \textbf{56.69} & \textbf{59.36} & \textbf{53.26}\\
    \bottomrule
\end{tabular}
\caption{The relevance discrimination and comparison capabilities of REAR and previous approaches. Generative LLMs struggle to determine the relevance degree of the given document, while our REAR overcomes it with the well-designed assessment module.
} 
\label{tab:relevance}
\end{table*}

%% file: sec/results.tex
\subsection{Main Results}
\label{subsec:mainres}
Table~\ref{tab:overall} shows the results of REAR and baselines on four open-domain QA tasks.

First, our REAR approach surpasses all the other baselines in QA performance. REAR not only performs well on the trained dataset, but also achieves good results on 
non-training datasets. This demonstrates that our precise signals for capturing relevance effectively guide the generation process. Thus, LLM can generate with good use of both parametric and external knowledge. 

Besides, the result shows the efficiency of data construction method, even without access to GPT APIs.
Self-RAG labels the relevance degree with GPT-4, while RobustLM and REAR utilize our proposed label machine and sampling method. The result indicates that our data construction strategy is effective and less costly. 

Third, generative LLMs struggle to determine the reliability degree of the given document, while our REAR overcomes it with the well-designed assessment module. As shown in Table~\ref{tab:relevance}, even the fine-tuned generative approaches~(RobustLM-training and Self-RAG) adequately discriminate relevance. In comparison, REAR significantly enhances this capability, highlighting its effectiveness in architectural design.

\subsection{Detailed Analysis}
\label{subsubsec:detailed}
In this part, we further present the analysis of the ablation study and impacts of retrieved documents. 

\subsubsection{Ablation Study}
\label{subsec:ablation}
We analyze how each of the proposed components affects final performance. Table ~\ref{tab:rar_comparison} shows the performance of our default method and its five variants in three aspects, including the architecture, training objectives and sampling strategy.

(1) \emph{w/o Assessment}: the variant without the integration of the rating module.
We utilize language generation to assess relevance degrees instead of the rank head. The document is selected based on the probability of generating judgmental statements. There is a notable drop in the comparison accuracy (see Hit@1 metric), similar shortfall is also observed in Self-RAG (Table~\ref{tab:overall}).
This demonstrates the effectiveness of our architectural design, 
which not only minimizes interference between language generation and relevance discrimination, but also facilitates the incorporation of various loss functions.

(2) \emph{w/o Consistency}: using the path-reliability strategy instead of the knowledge consistency strategy.
The path-reliability approach achieves higher Hit@1 rates, yet falls behind in EM and F1 scores compared to the knowledge-consistency strategy. 
The latter conducts a self-verification of outputs based on its generation ability, effectively integrating inherent knowledge in relevance assessment, which enhances the accuracy of RAG.

\begin{table}[t]
\centering
\small
\begin{tabular}{lccccc}
\toprule
\textbf{Methods} & \textbf{Aspect} & \textbf{Hit@1} & \textbf{EM} & \textbf{F1}    \\
\midrule
\textbf{REAR} & - & \uline{66.79} & \textbf{53.13} & \textbf{61.84}\\
\midrule
w/o $\text{Assessment}$ & Arch. & 13.80 & 38.14 & 47.44 \\
w/o $\text{Consistency}$ & Arch. & \textbf{67.48} & \uline{52.91} & \uline{61.49}\\
w/o $\text{Bi-granularity}$ & Obj. & 66.54 & 51.88 & 59.91 \\
w/o $\text{Noise-resistant}$ & Obj. & 49.25 & 25.54 & 33.05 \\
w/o $\text{Sampling}$ & Sam. & 61.99 & 49.00 & 53.62 \\
\bottomrule
\end{tabular}
\caption{Ablation study on our REAR. The ``aspect'' denotes the affected aspect. Arch., Obj. and Sam. denote architecture, training objective and sampling strategy.}
\label{tab:rar_comparison}
\end{table}

(3) \emph{w/o Bi-granularity}: the variant without bi-granularity fusion in relevance assessment training. We replace the bi-granularity loss with the coarse-grained loss function.
The results indicate that the fine-grained relevance training could enhance the LLMs in relevance comparison among documents, and result in better performance.

(4) \emph{w/o Noise-resistant}: the variant without noise-resistant training. We exclude the gold-noise data pairing, using the similar training construction approach of Self-RAG and RobustLM, with one document per query. We observe a notable decline, underscoring the effectiveness of noise-resistant training to enhance generation against irrelevant document interference.

(5) \emph{w/o Sampling}: the variant training with random hard negatives for training. We can observe a significant drop in relevance assessment capability, further illustrating the effectiveness of our method.

\subsubsection{Impact of Retrieved Documents} 

In this part, we further analyze the impact of retrieved documents in both single-document and multi-document settings.  

\paratitle{Single-Document Setting.}
We first examine the impact of external evidence in single document setting, where only the top first retrieved document is taken for reference. 
Table~\ref{tab:re-ir} shows the factual accuracy of different LLMs. We can see that both Self-RAG and REAR, after fine-tuning, perform well in relevant document utilization. However, REAR significantly outperforms other LLMs in generating accurate responses when the reference document is irrelevant, highlighting its robust resistance to interference from noisy documents.

\begin{table}[t]
\centering
\small
\begin{tabular}{@{}lcccc@{}}
\toprule
\multirow{2}{*}{\textbf{LLM}} & \multirow{2}{*}{\textbf{Settings}} & \textbf{Rel Doc} & \textbf{Irr Doc} & \textbf{Overall} \\
& & \scriptsize{(EM/Acc)} & \scriptsize{(EM/Acc)} & \scriptsize{(EM/Acc)} \\
\midrule
LLaMA2$_\text{7B}$ & $\text{4-shot}$ & 54.41 & \uline{6.40} & 30.53\\
LLaMA2$_\text{13B}$ & $\text{4-shot}$ & 53.36 & \uline{6.40} & 30.00\\
Mistral$_\text{7B}$ & $\text{4-shot}$ & 36.05 & 2.00 & 19.11\\
Baichuan2$_\text{7B}$ & 4-shot & 48.68 & 5.96 & 27.42\\
ChatGLM3$_\text{6B}$ & $\text{4-shot}$ & 46.97 & 2.12 & 24.65\\
Self-RAG$_\text{7B}$ & fine-tuned & \uline{73.48} & 6.23 & \uline{40.03}\\
REAR$_\text{7B}$ & fine-tuned & \textbf{73.84} & \textbf{20.09} & \textbf{46.79} \\
\bottomrule
\end{tabular}
\caption{Results of factual generation accuracy provided with top-1 retrieved documents on the test set of NQ. Categorized by performance when providing relevant (Rel) and irrelevant (Irr) documents.}
\label{tab:re-ir}
\end{table}

\begin{figure}[t]
    \centering
    \includegraphics[width=0.48\columnwidth]{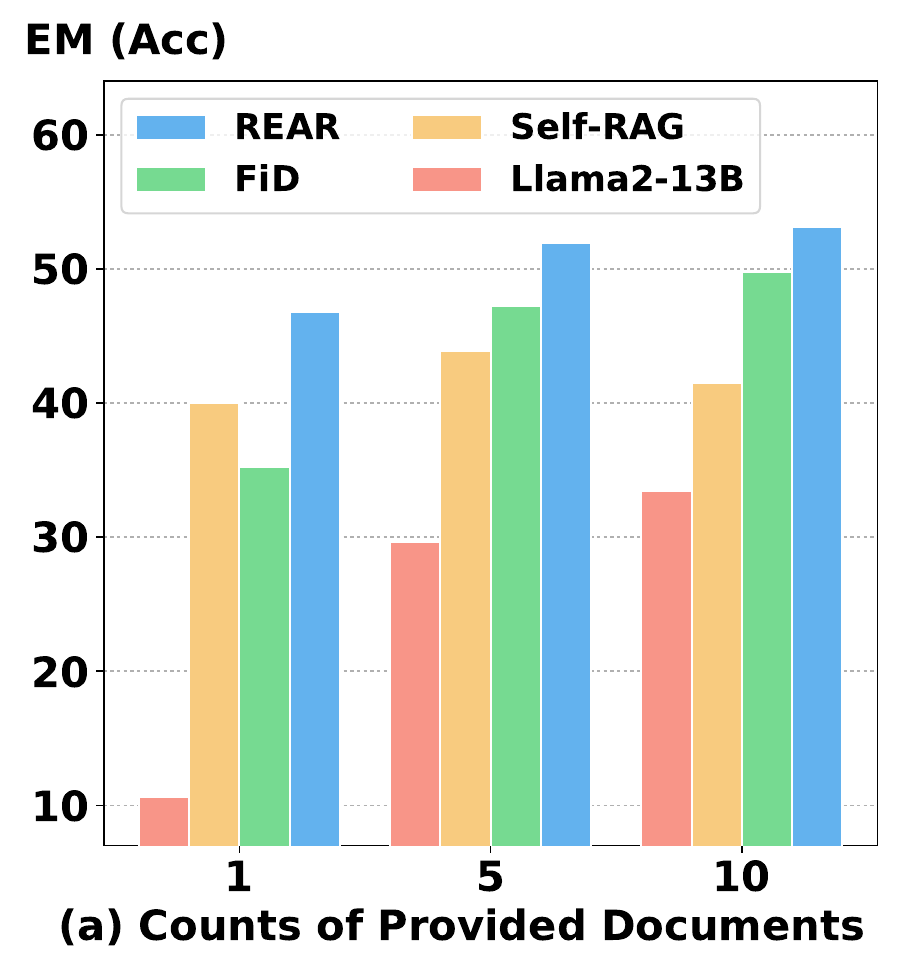} 
    \vspace{2pt} 
    \includegraphics[width=0.48\columnwidth]{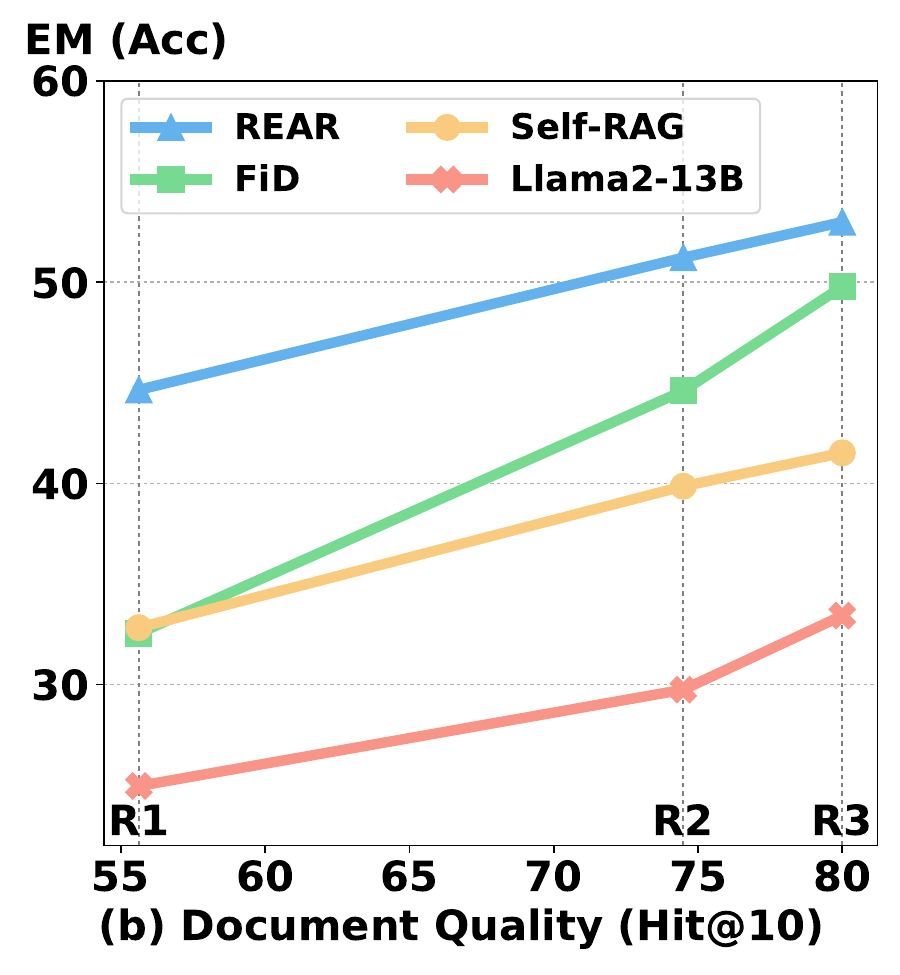} 
    \vspace{-5pt} 
    \caption{Results of RAG performance vary in overall document count and quality. The left one presents RAG performance with varying numbers of retrieved documents.
    The right one is the results of RAG with different retriever engines. R1, R2, and R3 represent BM25, Contriever-msmarco, and the FiD-distilled retriever, 
    \text{R1} < \text{R2} < \text{R3}~(Table~\ref{tab:nq-re} of the Appendix)
.}
    \captionsetup{skip=2pt}
    \label{fig:mainfig}
\end{figure}

\paratitle{Multi-Document Setting.}
In the second setting, we assume that  multiple retrieved documents can be used for reference. 
Specially, we mainly examine the impact of the \emph{total number} and \emph{relevance degree} of reference documents. For this purpose,  we vary the number of provided documents (Fig.~\ref{fig:mainfig}(a)) and the retriever's capabilities (Fig.~\ref{fig:mainfig}(b)). From Fig.~\ref{fig:mainfig}(a), we can see that our REAR approach performs well when provided with a single document (\ie the top retrieved one), while base models without fine-tuning suffer from significant degradation in this case. Furthermore, as shown in Fig.~\ref{fig:mainfig}(b), our approach is very robust to external retrievers of varied retrieval capacities. Especially,  when equipped with the weakest retriever BM25, it yields a large improvement over the other baselines, which further demonstrates that our approach can effectively perceive the relevance of external evidence for more suitable utilization.

%% file: sec/appendix.tex
\clearpage
\appendix
\section{Details on Fine-Gained Relevance Optimization}
\label{app:label}

\begin{figure}
\centering
\includegraphics[width=0.44\textwidth]{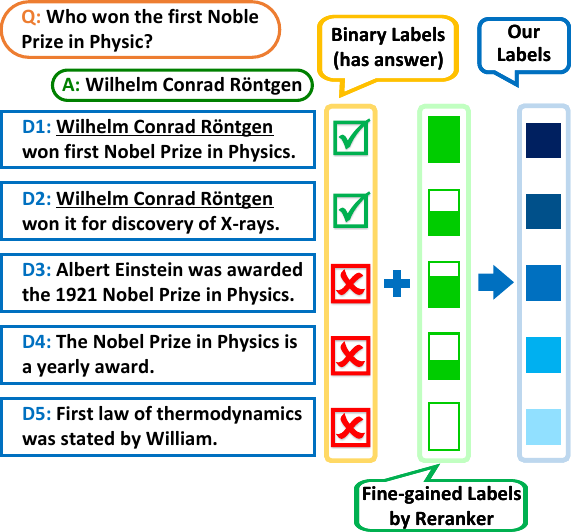}
    \caption{The illustration of different retrieved documents and different labeling metrics.}
    \label{fig:documents}
\end{figure}

We first illustrate why to design the fine-grained optimization for the assessment module.
Traditional annotation methods always use a binary labeling method~\cite{karpukhin2020dense}, which is based on the presence of an answer within a document. As shown in Fig.~\ref{fig:documents}, both D1 and D2 are labeled as ``\emph{relevant}''. However, while D1 allows for direct answer derivation, D2 requires additional external knowledge for induction. Training models solely on simple binary classification fails to distinguish the superiority of D1 over D2, potentially leading to inaccuracies in finer relevance judgments.

Previous work has achieved success in relevance assessment by distilling the ranking results from GPT-4~\cite{Sun2023IsCG}.
Inspired by it, we propose a less costly solution by labeling with small-scale, well-trained cross-encoder rerankers RocketQAv2~\cite{ren2021rocketqav2}. 
Despite the good relevance evaluation performance, it still may get wrong.
We adopt three strategies to reduce the negative impact of annotation errors on training.
Firstly, we design the sampling method (Eq.~\ref{eq:hard_neg}), which reduces the likelihood of potentially false negatives being sampled.
Besides, we linearly combine the binary label with cross-encoder scores. Thirdly, to mitigate noise from rerankers, we disregard differences smaller than 0.1 in fine-gained relevance training in Eq.~\ref{eq:fine}. These strategies enhance the quality of training data, which in turn improves the performance of REAR.

\section{Details on Dataset.}
\label{app:data}

We utilize four open-domain QA datasets, Natural Questions~(NQ)~\cite{kwiatkowski2019natural}, TriviaQA~\cite{joshi2017triviaqa}, WebQuestions~(WebQ)~\cite{berant2013semantic} and SQuAD~\cite{rajpurkar2016squad}). 

\begin{table}[ht]
\centering
\small
\begin{tabular}{lcccc}
\hline
\textbf{Dataset} & \textbf{NQ} & \textbf{TriviaQA} & \textbf{WebQ} & \textbf{SQuAD}\\
\hline
Num. Data & 3,610 & 11,313 & 2,032 & 10,570 \\
\hline
\end{tabular}
\caption{Dataset statistics of the test set.}
\label{tab:statistics}
\end{table}

\textbullet~\textbf{NQ:} a dataset designed to support comprehensive QA systems. It includes questions sourced from actual Google search queries. The corresponding answers are text spans within Wikipedia articles, meticulously identified by human annotators.

\textbullet~\textbf{TriviaQA:}  a compilation of trivia questions paired with answers, both of which were initially extracted from online sources.

\textbullet~\textbf{WQ:} constructed from questions proposed via the Google Suggest API, with answers being specific entities listed in Freebase.

\textbullet~\textbf{SQuAD:} a dataset for evaluating reading comprehension, and also is used for training and testing open-domain QA engines.

NQ is used for both training and inference, while the other three are only used for inference. We use the same split as previous work~\cite{karpukhin2020dense}. The training set of NQ contains 58,880 samples.

\section{Details on Document Collection}
\label{app:rere}

In this part, we introduce the retrievers we used to collect documents.
We employ task-specific retrievers to acquire the retrieved document. For inference, we utilize FiD-distilled retrievers~\cite{izacard2020distilling} for NQ and TriviaQA datasets. And we implement a strategy incorporating in-batch negatives and joint retriever-ranker training, starting from the Contriever-msmarco~\cite{izacard2021contriever} checkpoint for SQuAD and WQ datasets. 
The recall and MRR rates of the retrieved documents for inference are shown in Table.~\ref{tab:recall}.

\begin{table}[ht]
\centering
\small
\begin{tabular}{lcccc}
\hline
Metrics & NQ & TriviaQA & WQ & SQuAD \\
\hline
Hit@1 & 50.25 & 62.91 & 50.64 & 37.53 \\
Hit@10 & 80.00 & 81.78 & 75.89 & 68.51 \\
\hline
\end{tabular}
\caption{Retrievers we use for testing Hit rates (Recall rates) across datasets on the test sets.}
\label{tab:recall}
\end{table}

\begin{table}[ht]
\centering
\small
\begin{tabular}{lccc}
\hline
\textbf{Metric} & \textbf{R1} & \textbf{R2} & \textbf{R3} \\
\hline
Hit@10 & 55.62 & 74.49 & 80.00 \\
MRR@10 & 32.35 & 51.45 & 60.32 \\
\hline
\end{tabular}
\caption{Performance of three retrievers on the NQ test set. Hit@10 measures the percentage of correct answers within the top 10 results, indicating the precision of the retriever. MRR@10 (Mean Reciprocal Rank at 10) calculates the average of the reciprocal ranks of the first correct answer within the top 10 results, reflecting the effectiveness and rank of correct answers by the system. R1, R2 and R3 denote BM25~\cite{robertson1995okapi}, Contriever-msmarco~\cite{izacard2021contriever} and the dense retriever~\cite{izacard2020distilling} trained by distilling attention scores of FiD reader~\cite{izacard2020leveraging}}
\label{tab:nq-re}
\end{table}

\section{Details on Implementation}
\label{app:train-inf}
Following the previous work~\cite{asai2023self}, we apply joint optimization combining relevance assessment and relevance-guided generation, as specified in Eq.~\ref{eq:loss}. The training utilizes a learning rate of 1e-6, a warm-up ratio of 0.03, a batch size of 64 and a cosine scheduler for 1 epoch. Our experiments leverage the computational power of 8 NVIDIA Tesla A100 GPUs, each with 40G of memory.

\begin{figure}[t!]
    \centering
    \begin{tikzpicture}
        \node[draw, rounded corners, inner sep=10pt, align=left, text width=0.44\textwidth] {
            \textbf{Knowledge:}\\
            \{retrieved document 1\}\\
            \{retrieved document 2\}\\
            ......\\
            \{retrieved document 9\}\\[1ex]
            Answer the following question with a very short phrase, such as ``1998'', ``May 16th, 1931'', or ``James Bond'', to meet the criteria of exact match datasets.\\[1ex]
            \textbf{Question:} \{question\}\\[1ex]
            \textbf{Answer:}
        };
    \end{tikzpicture}
    \caption{Prompts for ``\emph{direct RAG QA}''.}
    \label{fig:instruction-format}
\end{figure}

\section{Details on Baselines}
\label{app:baseline}
In this part, we detail the prompt design and inference details for baselines. 
For the prompt-based inference, we utilize the instruction-tuned open-source models obtained from Hugging Face.
Following the existing work~\cite{ren2024bases, tang2024unleashing, asai2023self}, we use the greedy decoding strategy for inference.
The specific instruction formats used in our tests are illustrated in Fig.~\ref{fig:instruction-format} and Fig.~\ref{fig:instruction-format-judgment}.

\begin{figure}[ht]
    \centering
    \begin{tikzpicture}
        \node[draw, rounded corners, inner sep=10pt, align=left, text width=0.44\textwidth] {
            {Given a passage and a query, predict whether the passage includes an answer to the query by producing either `Yes' or `No'. And then answer with the given passage if `yes', or answer with your external knowledge if `No'.}\\[1ex]
            \textbf{Passage:} \{positive document example 1\}.\\
            \textbf{Query:} \{question 1\}\\
            \textbf{Judge:} Yes.\\
            \textbf{Answer:} \{answer 1\}\\[1ex]
            \textbf{Passage:} \{negative document example 2\}.\\
            \textbf{Query:} \{question 2\}\\
            \textbf{Judge:} No.\\
            \textbf{Answer:} \{answer 2\}\\[1ex]
            \textit{Other 2 examples}...\\[1ex]
            \textbf{Passage:} \{one of the retrieved document\}.\\
            \textbf{Query:} \{question\}\\
            \textbf{Judge:} (calculate the difference in log perplexity for ``Yes'' and ``No'' and fill accordingly).\\
            \textbf{Answer:}
        };
    \end{tikzpicture}
    \caption{Prompts for ``\emph{RobustLM based prompting}''.}
    \label{fig:instruction-format-judgment}
\end{figure}